\documentclass[10pt,twocolumn,letterpaper]{article}

\usepackage{iccv}              

\definecolor{iccvblue}{rgb}{0.21,0.49,0.74}
\usepackage[pagebackref,breaklinks,colorlinks,allcolors=iccvblue]{hyperref}
\usepackage{graphicx}
\usepackage{colortbl}
\usepackage{amsmath}
\usepackage{multirow}
\usepackage{multicol}
\usepackage{amsfonts}       
\usepackage{nicefrac}       
\usepackage{microtype}      
\usepackage{xcolor}         
\definecolor{mygray}{RGB}{230,230,230}

\def\paperID{7682} 
\def\confName{ICCV}
\def\confYear{2025}

\title{Aligning Effective Tokens with Video Anomaly in Large Language Models}

\author{Yingxian Chen \and Jiahui Liu \and Ruidi Fan \and Yanwei Li \and Chirui Chang \and Shizhen Zhao \and Wilton W.T. Fok \and Xiaojuan Qi \and Yik$-$Chung WU}

\author{Yingxian Chen$^{1*}$~~ Jiahui Liu$^{1*}$~~ Ruidi Fan$^1$~~ Yanwei Li$^2$~~ Chirui Chang$^1$~~ Shizhen Zhao$^1$~~\\
Wilton W.T. Fok$^1$~~ Xiaojuan Qi$^1$\footnotemark[2]~~ Yik-Chung Wu$^1$\footnotemark[2]~~ \\
$^1$The University of Hong Kong~~~ $^2$The Chinese University of Hong Kong\\
\hspace{-12pt}\texttt{\small \{chenyx, liujh, xjqi, ycwu\}@eee.hku.hk}\\
$^{*}$equal contributions~~~ \footnotemark[2]~corresponding authors
}

\begin{document}
\maketitle
\begin{abstract} Understanding abnormal events in videos is a vital and challenging task that has garnered significant attention in a wide range of applications. 
Although current video understanding Multi-modal Large Language Models (MLLMs) are capable of analyzing general videos, they often struggle to handle anomalies due to the spatial and temporal sparsity of abnormal events, where the redundant information always leads to suboptimal outcomes.
To address these challenges, exploiting the representation and generalization capabilities of Vison Language Models (VLMs) and Large Language Models (LLMs), we propose VA-GPT, a novel MLLM designed for summarizing and localizing abnormal events in various videos. 
Our approach efficiently aligns effective tokens between visual encoders and LLMs through two key proposed modules: Spatial Effective Token Selection (SETS) and Temporal Effective Token Generation (TETG). 
These modules enable our model to effectively capture and analyze both spatial and temporal information associated with abnormal events, resulting in more accurate responses and interactions. 
Furthermore, we construct an instruction-following dataset specifically for fine-tuning video-anomaly-aware MLLMs, and introduce a cross-domain evaluation benchmark based on XD-Violence dataset.
Our proposed method outperforms existing state-of-the-art methods on various benchmarks. 
\end{abstract}    
\section{Introduction}
\label{sec:intro}

Detecting and summarizing abnormal events in videos is critical and challenging, and it has garnered considerable attention across multiple research domains and real-world applications, such as security monitoring, video analysis, and crime detection.

Although many traditional methods~\cite{un-mem1,un-amcm,un-hf2,un-mnad,motion-aware,motion_amc,cleaning-label} have been widely explored for video anomaly detection, they exhibit substantial limitations in their effectiveness~\cite{claws,DMU,CLIP-TSA,wu2024OVVAD,weakly-ldf,weakly-wsvad}. These limitations manifest in two aspects: 1) Traditional video anomaly detection methods~\cite{MIST,motion-aware,RTFM,Not-only-look,LCTR,MGFN} essentially approach the task as a closed-set detection and classification problem, inherently limiting their ability to achieve a comprehensive understanding and interpretation of anomalies; 2) These methods~\cite{wu2023vadclip,yang2024textprompt,challen-1,challen-2,challen-3,challen-4} are restricted by a limited vocabulary, making it difficult for them to handle unseen or novel situations effectively.


\begin{figure}[t]
\centering
\vspace{0.1cm}
\includegraphics[width=0.99\linewidth]{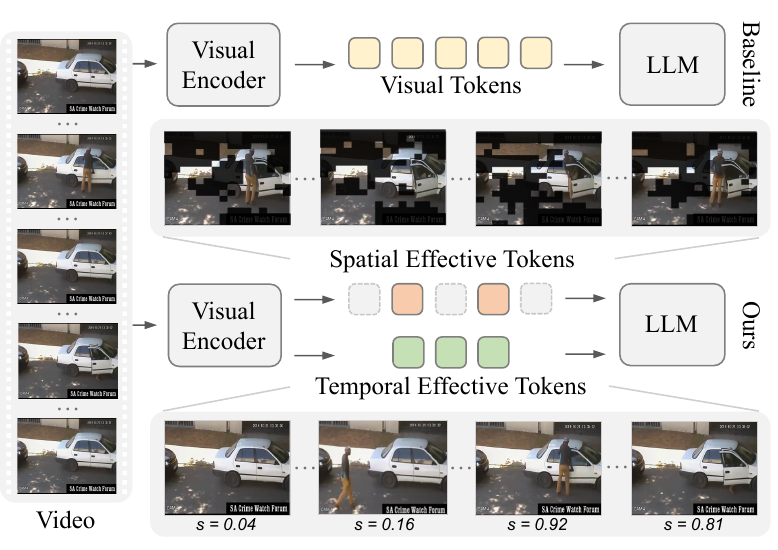}
\vspace{0.1cm}
\caption{Baseline video understanding MLLM feeds forward every visual token (yellow squares) equally to participate in fine-tuning and inference (top row). Different from it, our method focuses on the effective area (unobstructed area in medium video frames) in each frame and select the Spatial Effective Tokens (orange squares) for the LLM (see Section~\ref{subsec:SETS}) (filtered tokens are shown as gray squares). At the same time, we generate anomaly-aware Temporal Effective Tokens (green squares) (see Section~\ref{subsec:TETG}) based on the assigned anomaly scores (denoted as \emph{s}) of each frame from a pre-trained classifier for better temporal localization of anomalies.}
\label{fig:OVERVIEW}
\vspace{-0.4cm}
\end{figure}

Recent advancements~\cite{llava,li2023blip2,chatgpt,gpt4,llama, liu2024can} in Vision Language Models (VLMs) and Large Language Models (LLMs) have demonstrated remarkable capabilities in scene understanding and comprehensive analysis. Multimodal Large Language Models (MLLMs), particularly those designed for video understanding~\cite{videollama,videochat,valley,li2023llamavid,videochatgpt,mplug}, have achieved significant progress in general video analysis tasks. However, while these models exhibit strong performance in general video understanding, they fall short in accurately detecting and interpreting anomalies.

For mitigating the above challenges, some works~\cite{Holmes-VAU,tang2024hawk,zhao2024hawkeye,du2024ECVA,yang2024followrulesreasoningvideo} proposed anomaly-aware video MLLMs to better understand the anomalies in videos. Although these models work well for detecting obvious abnormal events, such as fighting or fire, they typically struggle to effectively align abnormal regions with relevant captions which requires addressing spatial redundancy, and accurately identifying abnormal time intervals by mitigating temporal redundancy. This is because these methods treat all latent tokens with equal priority across spatial and temporal dimensions. This leads to performance degradation caused by redundant tokens unrelated to anomalies. However, in most cases, only small regions within a few frames contain the essential information help to identify an anomaly (as shown in Figure~\ref{fig:OVERVIEW}). 
Thus, we explore: \emph{How can multimodal architectures evolve selective token generation and processing mechanisms to dynamically prioritize anomaly-salient information while preserving comprehensive scene understanding capabilities?} 


To address the aforementioned issues, 
we propose a new model named \textbf{VA-GPT} for analyzing various \textbf{V}ideos for \textbf{A}bnormal events by aligning effective and accurate tokens with LLMs across both \emph{spatial} and \emph{temporal} dimensions.
VA-GPT integrates two key components to identify effective visual tokens for alignment while eliminating redundant tokens that could hinder anomaly analysis and distract model from extracting useful information: 1) we develop the Spatial Effective Token Selection (SETS) module for identifying tokens corresponding to regions with challenges for aligning them with LLMs, while filtering out tokens associated with minor dynamics to remove redundancy. This is because we find that abnormal events often result in different visual changes and variations in local areas (see Figure~\ref{fig:OVERVIEW}); and 2) 
we propose the Temporal Effective Token Generation (TETG) module which employs a lightweight pre-trained classifier to assign a confidence score to each frame indicating the possibility of containing abnormal events. Then TETG generates efficient tokens with prior knowledge of the temporal information of abnormal events directly in the language space as additional input to the LLMs, effectively enhancing the model's temporal reasoning and understanding abilities about abnormal events.

Furthermore, beyond conventional benchmarks (in-domain benchmark), we establish a new cross-domain evaluation protocol that systematically evaluates model robustness with domain shifts. Based on a novel video dataset, XD-Violence~\cite{Not-only-look}, we design comprehensive QAs about abnormal events which include different visual contents from our training data and integrate it as a new cross-domain benchmark. Meanwhile, we design temporal-information-oriented QAs on both in- and cross- domain benckmarks for evaluating temporal localization abilities.
Comprehensive experiments demonstrate VA-GPT's superiority, achieving state-of-the-art performance in both in-domain anomaly localization and cross-domain generalization scenarios. 

The main contributions are summarized as follows:

\begin{itemize}
\item We propose VA-GPT, a video-anomaly-aware MLLM for detecting and summarizing anomalies in various videos, which introduces the MLLM to the specific domain of video anomaly understanding.

\item  We introduce the SETS and TETG, which enable our MLLM to effectively capture both spatial and temporal information in video sequences, resulting in accurate understanding and localization of abnormal events. Meanwhile, we propose a new instruct-following dataset for video anomaly analysis and a comprehensive cross-domain evaluation benchmark for better evaluating the generalization abilities of MLLMs on video anomalies.

\item Our extensive experiments demonstrate that our method outperforms existing state-of-the-art methods in various benchmarks, highlighting its effectiveness and potential for practical applications in video anomaly understanding.
\end{itemize}
\section{Related Work}

\begin{figure*}[t]
\centering
\includegraphics[width=1.0\linewidth]{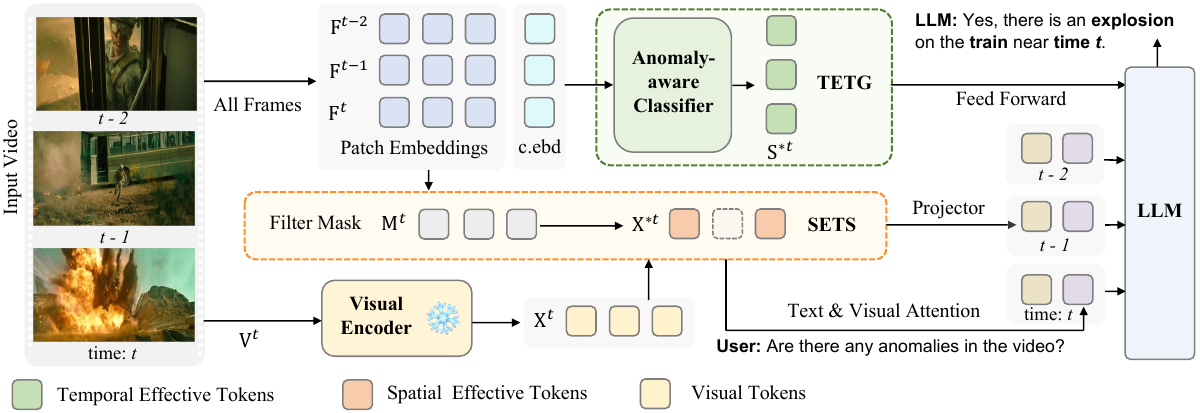}
\vspace{-0.4cm}
\caption{Detailed illustration of our proposed model. When a video is fed into the model, patch embeddings \includegraphics[width=2mm]{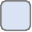} and class embeddings (c.ebd) \includegraphics[width=2mm]{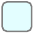} are extracted from all frames. 1) Based on the difference in patch embeddings between current frame and its neighbour frame, we can get a filter mask \includegraphics[width=2mm]{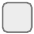} to filter out unimportant visual tokens (dashed square \includegraphics[width=2mm]{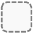}) from current frame's visual tokens \includegraphics[width=2mm]{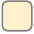}, thereby selecting Spatial Effective Tokens \includegraphics[width=2mm]{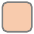} that are compressed with a projector with pooling into aligned content token \includegraphics[width=2mm]{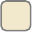} for each frame, meanwhile take attention with text input from users for resulting aligned context token \includegraphics[width=2mm]{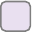} for each frame; 2) Based on class embeddings (c.ebd) \includegraphics[width=2mm]{img/icons/2.pdf} of all frames, we use a pre-trained Anomaly-aware Classifier to localize the time period of abnormal events, thereby generating Temporal Effective Tokens \includegraphics[width=2mm]{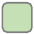} to feed forward into the LLM. All of the resulting aligned tokens \includegraphics[height=2mm]{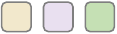} are fed into the LLM for reasoning and inference of the whole model.
}
\vspace{0cm}
\label{fig:OVERVIEW2}
\end{figure*}

\subsection{Large Language Models (LLMs)}
The domain of Natural Language Processing (NLP) has experienced significant progress, particularly with the emergence of Large Language Models (LLMs). The introduction of the Transformer architecture~\cite{trans-survey,trans1,trans2} is a critical turning point, followed by other influential language models~\cite{llms1,trans2,llms3,llms4} that exhibited remarkable proficiency. Generative Pre-trained Transformers (GPT)~\cite{gpt} brought about a revolution in NLP by employing auto-regressive prediction, establishing itself as a powerful language modeling approach. More recent groundbreaking contributions, such as ChatGPT~\cite{chatgpt}, GPT-4~\cite{gpt4}, LaMDA~\cite{Lamda} and LLaMA~\cite{llama}, have expanded the horizon even further. These models, trained on extensive textual data, display extraordinary performance in intricate linguistic tasks.

\subsection{Vision Language Models (VLMs)}

Progress in the fields of computer vision and natural language processing has given rise to the development of vision-language models (VLMs)~\cite{radford2021learning, jia2021scaling, yao2021filip, gao2022pyramidclip, wang2023visionllm, llava, lin2024moellava}. These models combine visual and linguistic systems to facilitate cross-modal comprehension and reasoning. Notable examples include CLIP ~\cite{radford2021learning} which pairs BERT~\cite{llms3} with ViT~\cite{vit}; BLIP-2~\cite{li2023blip2} incorporating Vision Transformer features into Flan-T5~\cite{chung2022scaling}; MiniGPT4~\cite{zhu2023minigpt4}, connecting BLIP-2 with Vicuna~\cite{li2023blip2,peng2023instruction}; PandaGPT~\cite{su2023pandagpt} , bridging ImageBind~\cite{girdhar2023imagebind} with Vicuna. These models excel in tasks like image classification, captioning, and object detection~\cite{vlm_imageclassification,vlm_object,su2023pandagpt}. Recent developments in vision-language models have extended into video processing with models like Video-Chat~\cite{videochat}, Video-ChatGPT~\cite{videochatgpt}, Otter~\cite{otter}, Valley~\cite{valley}, mPLUG~\cite{mplug}, Video-LLaMA~\cite{videollama}, and LLaMA-VID~\cite{li2023llamavid}. These systems enable interactive video querying, enhance comprehension through audio-visual-text alignment, and support comprehensive video analysis. In this paper, we leverage VLMs and LLMs to develop a novel approach for video anomaly understanding.


\subsection{Video Anomaly Understanding (VAU)}

Annotating surveillance video frames is labor-intensive, prompting researchers to explore alternatives: one-class learning~\cite{one-class}, unsupervised anomaly detection without annotations~\cite{un-amcm,un-cdda,un-fsad,un-hf2,un-mem1,un-mnad,un-oada}, or weakly supervised methods using only video-level annotations~\cite{wsstad,MIST,RTFM,LCTR,MSL,IBL,MGFN,weakly-wsvad,claws}. In one-class learning, Luo et al. developed a ConvLSTM network for normal segment learning~\cite{un-rhc}. Several researchers employed Auto-Encoder networks to reconstruct normal frame features~\cite{un-sta,un-oada,un-cdda}, while others implemented memory mechanisms~\cite{un-mem1,un-amcm,un-hf2,un-mnad} or meta-learning~\cite{un-fsad} to enhance generalizability. For weakly supervised learning, Tian~\cite{RTFM} sed multiple-instance learning to localize anomalous clips. Zhong et al. uutilized graph convolution networks, though with limited generalization capability. ~\cite{GCLNC}. To address this, Ramachandra et al. developed a Siamese network for normal feature learning. Wan et al. and Zaheer et al. ~\cite{weakly-wsvad,claws}proposed clustering-based frameworks for anomaly identification. Recent studies have introduced new architectures for spatial-temporal feature ensemble learning~\cite{wsstad,MIST,RTFM,LCTR,MSL,IBL,MGFN}. However, these methods merely supply anomaly scores during inference, necessitating empirical threshold establishment on test sets to differentiate abnormal events. Recent research has begun exploring MLLMs to enhance models' capabilities in identifying and describing anomalies~\cite{Holmes-VAU,tang2024hawk,zhao2024hawkeye,du2024ECVA,yang2024followrulesreasoningvideo}. 



\section{Method}


\subsection{Overview}

\textbf{Task.} Video anomaly understanding MLLMs aims to determine whether an input video contains abnormal events, meanwhile describing and interacting with the temporal localization and the entire process of the detected abnormal events (if has). We train the model with an instruct-following dataset based on abnormal videos~\cite{ucfcrime}, so that the model can better align the tokens between visual encoders and LLMs for presenting and generalizing information about abnormal events.

\noindent \textbf{Pipeline.} Considering the video understanding MLLM framework as shown in Figure~\ref{fig:OVERVIEW2}, taking a video (contains $T$ frames) as input, a frozen ViT-based~\cite{vit} visual encoder (CLIP~\cite{sanghi2021clip}) extracts visual tokens $\textbf{X}^t$ from each video frame $\textbf{V}^t (t=1,...,T)$. For $\textbf{X}^t=\left\{\textbf{x}^{t}_{i}\right\}_{i=1,...,N}$ of the current frame, there are $N$ visual tokens corresponding to equal amounts of image patches. Modality alignment converts the processed visual tokens $\textbf{X}^t$ into the semantic space of LLMs. At the same time, text prompts are processed and encoded as text tokens into the same semantic space and serve as a part of input to LLMs. Our key design on models consists of (1) selecting Spatial Effective Tokens $\textbf{X}^{*t}$ (SET) from $\textbf{X}^{t}$ for each frame participating in fine-tuning and inference instead of $\textbf{X}^{t}$ (see Section~\ref{subsec:SETS}); 
and (2) generating Temporal Effective Tokens $\textbf{S}^{*t}$ (TET) as anomaly-aware temporal priors, participating in inference to facilitate the temporal localization of abnormal events for LLMs (see Section~\ref{subsec:TETG}). In addition, we produce high-quality instruct-following data on abnormal videos and develop a training strategy for it to maximize the effectiveness of our proposed method (see Section~\ref{subsec:training}).

\subsection{Spatial Effective Token Selection (SETS)}
\label{subsec:SETS}

In classical video classification tasks, context and relationships are critical. However, in our MLLM setting, beyond leveraging contextual information, the most crucial problem is aligning the visual and language modalities. Therefore, the key aspect of our design is to extract useful information for effectively aligning visual tokens with the LLM. Since text captions primarily describe anomaly events, which occupy only a small portion of the entire video, aligning all visual patterns with text tokens would be unreasonable and computationally heavy. Thus, we are the first to propose a novel token selection method SETS to achieve efficient and effective alignment.

\noindent \textbf{Inter-frame Difference.} For a video, we believe that areas with large changes in adjacent frames are more worthy of attention. As illustrated in Figure~\ref{fig:OVERVIEW2}, for each frame $\textbf{V}^t$ of the video, we can regard its previous frame $\textbf{V}^{t-1}$ as the reference frame for investigating the difference between current timestamp and previous timestamp. Employing DINOv2~\cite{dinov2} as the feature extractor, denoted as $\mathcal{F}_{\text{E}}$, we can extract patch embeddings:
\begin{equation}
\textbf{F}^t, \textbf{F}^{t-1} = \mathcal{F}_{\text{E}}(\textbf{V}^t), \mathcal{F}_{\text{E}}(\textbf{V}^{t-1}),
\label{eq:extract}
\end{equation}
where $\textbf{F}^t,\textbf{F}^{t-1} \in \mathbb{R}^{N\times C}$ are the extracted embeddings ($N$ indicates the number of image patches and $C$ indicates the channels). Thanks to the distinction and stability of the extracted features, we calculate their patch-wise distances as the inter-frame difference map of the current frame:
\begin{equation}
\textbf{D}^t = dis(\textbf{F}^t, \textbf{F}^{t-1}),
\label{eq:dis_map}
\end{equation}
where $dis(\cdot)$ indicates Manhattan distance~\cite{Manhattan} and $\textbf{D}^t \in \mathbb{R}^{N}$ indicates the distances between corresponding patch pairs in neighbour frames.

\noindent \textbf{Select Spatial Effective Tokens.} According to the inter-frame difference map $\textbf{D}^t$, we can set up a vector $\textbf{M}^t=[m^{t}_1, m^{t}_2, ..., m^{t}_N]$ to record the difference of each patch, where top $K$ ratio of elements with the largest distance are assigned with the value of 1, and the rest are assigned as 0. Thus we get a mask for filtering and updating the visual tokens as:
\begin{equation}
\textbf{X}^{*t}=\left\{\left.\textbf{x}^{t}_i\middle|\right.m^{t}_i=1, m^{t}_i\in\textbf{M}^t\right\},
\label{eq:filter}
\end{equation}
where $\textbf{X}^{*t}$ contains the selected Spatial Effective Tokens (SET) which are fed into subsequent processing instead of $\textbf{X}^{t}$ as shown in Figure~\ref{fig:OVERVIEW2}. SET can efficiently isolate the regions highly related to the abnormal events to participate in both fine-tuning and inference.

\begin{table*}
\centering
\renewcommand{\arraystretch}{1.0}
\begin{center}
\resizebox{0.99\linewidth}{!}{
\begin{tabular}{lccccc}
\toprule
\multirow{2}{*}{\textbf{Method}} & \multirow{2}{*}{\textbf{LLM}} & \multicolumn{2}{c}{\textbf{In-domain}} &  \multicolumn{2}{c}{\textbf{Cross-domain}}  \\
 & & \textbf{Total Acc.(\%)} & \textbf{Temporal Acc.(\%)} & \textbf{Total Acc.(\%)} & \textbf{Temporal Acc.(\%)} \\
\midrule
Video-Chat~\cite{videochat} & Vicuna-7B & 22.41 & - & - & -\\
Video-ChatGPT~\cite{videochatgpt} & Vicuna-7B & 24.13 & 28.51 & 24.00 & 29.10\\
Otter~\cite{otter} & LLaMa-7B & 22.41 & 22.17 & 25.20 & 23.80 \\
Valley~\cite{valley} & Vicuna-7B & 20.34 & 14.48 & 21.00 & 20.20\\
mPLUG~\cite{mplug} & LLaMa-7B & 22.76 & - & - & -\\
Video-LLaMA2~\cite{cheng2024videollama} & Vicuna-7B & 21.38 & 26.62 & 24.20 & 23.00\\
Hawkeye~\cite{zhao2024hawkeye} & LLaVA-7B & 28.60 & 30.00 & 25.30 & 28.50\\
LLaMA-VID~\cite{li2023llamavid}    & Vicuna-7B & 14.83 & 26.70 & 18.80 & 23.60\\
\cellcolor{mygray}VA-GPT (\emph{Ours}) & \cellcolor{mygray}Vicuna-7B & \cellcolor{mygray}\textbf{30.69} & \cellcolor{mygray}\textbf{35.00} & \cellcolor{mygray}\textbf{26.20} &\cellcolor{mygray}\textbf{31.02} \\
\bottomrule
\end{tabular}}
\vspace{-0.1cm}
\caption{Comparing on in-domain (UCF-Crime~\cite{ucfcrime}) and the proposed cross-domain (XD-Violence~\cite{Not-only-look}) benchmarks, our method significantly outperforms other models and achieve the state-of-the-art performance (accuracy is detonated as Acc.) on anomaly video understanding and temporal localization with LLMs (Best results are shown in \textbf{bold}).}
\vspace{-0.4cm}
\label{table:videobenchresults}
\end{center}
\end{table*}
\setlength{\tabcolsep}{1pt}

\subsection{Temporal Effective Token Generation (TETG)}
\label{subsec:TETG}

\textbf{Anomaly-aware Classifier.} We design a simple but effective MLP $\mathcal{F}_{\text{A}}$ for learning whether each frame is related to an abnormal event. For the class embeddings (denoted as $\textbf{z}$) extracted from the feature encoder, we can split them based on training video caption into normal and anomaly embeddings, denoted as $\textbf{z}^{\text{n}}$ and $\textbf{z}^{\text{a}}$, respectively. Thus we can optimize $\mathcal{F}_{\text{A}}$ using a binary classification loss:
\begin{equation}
\resizebox{0.91\hsize}{!}{
$\mathcal{L} =\mathbb{E} _{\textbf{z}\sim\textbf{z}^{\text{n}}}\left[-\log\frac{1}{1 + \exp^{-\mathcal{F}_\text{A}(\textbf{z})}}\right]+\mathbb{E} _{\textbf{z}\sim\textbf{z}^{\text{a}}}\left[-\log\frac{\exp^{-\mathcal{F}_\text{A}(\textbf{z})}}{1+\exp^{-\mathcal{F}_\text{A}(\textbf{z})}} \right].
$}
\label{eq:classifer}
\end{equation}
The anomaly-aware classifier predicts whether each frame is related to anomalies in a video, which can bring additional important prior knowledge to LLMs at a very low cost to facilitate its inference.

\noindent \textbf{Generate Temporal Effective Tokens.} Since the information drawn from the anomaly-aware classifier is explicit, we can easily project it to LLMs' text token space through natural languages. Based on the prediction results of the anomaly-aware classifier, we select the first and last frames' timestamps with high confidence of containing abnormal events, denoted as \texttt{<a-start>} and \texttt{<a-end>}, respectively. Then we tokenize them with a template as: \texttt{"Known common crime types are: `Shooting',`Arson',`Arrest', ... There is one of the crime types occurring from <a-start> to <a-end>"}, resulting Temporal Effective Tokens (TET) in the text token space of LLMs. During inference, with the well-trained lightweight anomaly-aware classifier, TET is used as an additional input to participate in the forward process of the LLM to provide prior knowledge about abnormal events temporally (as shown in Figure~\ref{fig:OVERVIEW2}).

\subsection{Training Strategy}
\label{subsec:training}

For modality alignment and instruction tuning, we follow the baseline~\cite{li2023llamavid} to ensure visual features are well aligned with the language space. In this work, the training strategy can be divided into two stages: 1) Stage One: Fine-tuning with anomaly video data, and 2) Stage Two: Aligning Spatial Effective Tokens with LLMs.


\noindent \textbf{Fine-tuning with Video Anomaly Data.} For enhancing the abnormal scene understanding of LLMs, we construct the UCF-crime~\cite{yuan2023surveillance} Question-Answer pairs for fine-tuning. We also mix different instruction pairs from various sources~\cite{lin2023videollava}, including text conversations, single/multi-turn visual Question-Answer pairs, and video Question-Answer pairs. Different formats for text, image, and video inputs are adopted, and the image token is randomly inserted at the beginning or end of user input during training. All modules, except the frozen visual encoder, are optimized in this stage. After fine-tuning, the LLM has a preferred perception of anomalies, thus ensuring the effectiveness of the temporal effective tokens (see Section~\ref{subsec:TETG}) during inference. More dataset details are illustrated in Section~\ref{Experiments}.

\noindent \textbf{Aligning Spatial Effective Tokens with LLMs.} For abnormal video scenes, most areas would not be aligned with languages. Therefore, we implement an additional fine-tuning step. This step involves utilizing Spatial Effective Tokens (see Section~\ref{subsec:SETS}) derived from each video frame within the UCF-Crime dataset. By incorporating these tokens, we aim to provide the model with a more refined understanding of the spatial context of anomalies. It also brings efficient optimization, and the alignment here is only designed for very short-term fine-tuning, which can greatly improve the model's ability to detect and understand anomalies.

\begin{figure*}[t]
\centering
\includegraphics[width=1\linewidth]{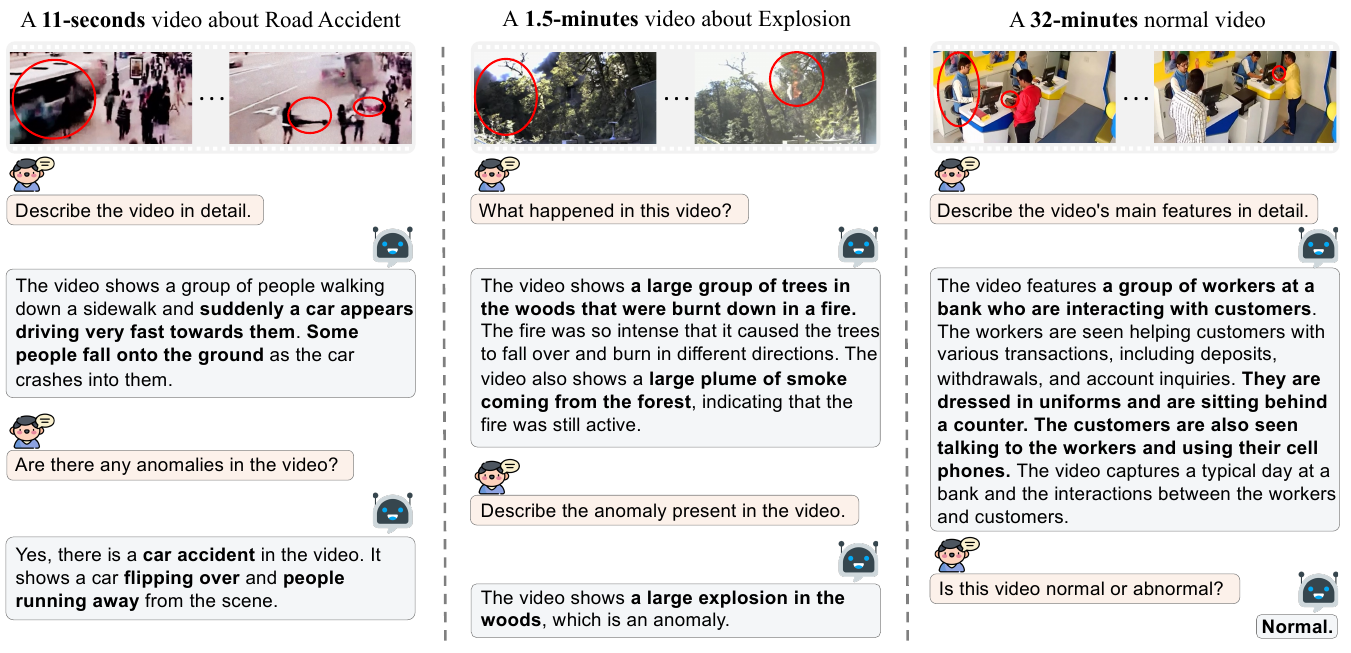}
\vspace{-0.6cm}
\caption{Qualitative results in Question-Answer diagrams, the red circles in the figures correspond to the \textbf{bold} text in the answers. From short video of only a dozen seconds to medium video of longer than one minute and long video of about half an hour, our model can reason well and understand the content.} 
\label{fig:interaction}

\end{figure*} 
\section{Experiments}
\label{Experiments}

\begin{table}[t]
\small
\centering
\renewcommand{\arraystretch}{1.0}
\resizebox{0.99\linewidth}{!}{ 
\begin{tabular}{l>{\centering\arraybackslash}p{2.1cm}>{\centering\arraybackslash}p{2.1cm}>{\centering\arraybackslash}p{2.1cm}}
    \toprule
    \textbf{Method} & \textbf{Description} & \textbf{Causes} & \textbf{Effect} \\
    \midrule
    A-Guardian~\cite{du2024ECVA} & 79.65 & 58.92 &  50.64 \\
    \cellcolor{mygray}\emph{Ours} & \cellcolor{mygray}80.83 & \cellcolor{mygray}59.55 & \cellcolor{mygray}51.08 \\
    \bottomrule
\end{tabular}}
\vspace{-0.2cm}
\caption{We evaluate our method on the MMEval~\cite{du2024ECVA} benchmark and show that our method outperforms the related previous method in different aspects.}
\vspace{-0.5cm}
\label{table:mmeval}
\end{table}

\textbf{Datasets.} We fine-tune our model on our proposed instruct-following format~\cite{llava} training dataset including 4077 videos and 7730 images based on UCF-Crime~\cite{ucfcrime} dataset. We evaluate the models on two anomaly video understanding benchmarks: UCF-Crime~\cite{ucfcrime} for in-domain evaluation and a proposed benchmark designed based on XD-Violence~\cite{Not-only-look} dataset for cross-domain evaluation, respectively. More details are shown in the supplementary.


\noindent \textbf{Benchmarks and Metrics.} To evaluate the ability to review videos and identify anomalies, we utilize a video anomaly understanding evaluation from Video-Bench~\cite{ning2023videobench} to assess the temporal comprehensive ability, which contains nature language based Question-Answer pairs from UCF-Crime~\cite{ucfcrime} dataset. Meanwhile, in order to evaluate the model's cross-domain video anomaly understanding ability, we contribute Question-Answer pairs as an extra benchmark based on the XD-Violence dataset. These Question-Answer pairs encompass four options, with each presenting the anomaly category and the respective time intervals during which the anomalies transpire. For each benchmark, we design different sets of questions for two evaluations: one is an overall evaluation of abnormal event detection and understanding, and the other is a special evaluation focusing on temporal localization ability, measured by question answering accuracy (denoted as Total Acc. and Temporal Acc., respectively, higher is better).


\noindent \textbf{Implementation Details.} For network structure, we incorporate the pre-trained CLIP~\cite{sanghi2021clip} and DINOv2~\cite{dinov2} as the visual encoder and Qformer~\cite{dai2023instructblip} as the text decoder. We follow ~\cite{li2023llamavid} to freeze the encoder during the modality alignment and to optimize the trainable parameters using anomaly videos and instructions for instruction tuning. During the training process, our model utilizes PyTorch on four NVIDIA A100 GPUs. We employ the AdamW optimizer with a batch size of 64. The learning schedule is set to cosine decay, with a learning rate of 2e-5 and a total of 1 epoch.

\subsection{Main Results}

\begin{table*}[t]
\centering
\renewcommand{\arraystretch}{1.1}
\resizebox{0.99\linewidth}{!}{ 
\begin{tabular}{l>{\centering\arraybackslash}p{1.40cm}>{\centering\arraybackslash}p{1.40cm}>{\centering\arraybackslash}p{1.40cm}>{\centering\arraybackslash}p{1.40cm}>{\centering\arraybackslash}p{1.40cm}>{\centering\arraybackslash}p{1.40cm}>{\centering\arraybackslash}p{1.40cm}>{\centering\arraybackslash}p{1.40cm}>{\centering\arraybackslash}p{1.40cm}>{\centering\arraybackslash}p{1.40cm}>{\centering\arraybackslash}p{1.40cm}}
    \toprule
    \multirow{2}{*}{\textbf{Method}} & \multicolumn{4}{c}{\textbf{Baseline}} & \multicolumn{3}{c}{\textbf{Stage One Fine-tuning}} & \multicolumn{3}{c}{\textbf{Stage Two Fine-tuning}}\\
    \cmidrule(lr){2-5} \cmidrule(lr){6-8} \cmidrule(lr){9-11}
    & w/o Both & w.SETS & w.TETG & w.Both & w.SETS & w.TETG & w.Both & w.SETS & w.TETG & w.Both \\
    \midrule
    \textbf{Total Acc. (\%)}$\uparrow$ & 14.83 & 24.83  & 23.79  & 25.12 & 25.86 & 26.10 & 27.50 & 29.31 & 28.96 & \cellcolor{mygray}\textbf{30.69}\\
    \textbf{Temporal Acc. (\%)}$\uparrow$ & 26.70 & 27.20 & 27.76 & 28.81 & 29.68 & 30.02 & 30.77 & 31.60 & 33.58 & \cellcolor{mygray}\textbf{35.00}\\
    \bottomrule
\end{tabular}}
\vspace{-0.05cm}
\caption{Ablation studies on Spatial Effective Token Selection (SETS), Temporal Effective Tokens Generation (TETG), and progressive training strategies. At different model training stages, starting from the baseline (w/o fine-tuning, w/o both), we compare the performance of only using SETS (w.SETS), only using TETG (w.TETG), and using both (w.Both) on the UCF-Crime benchmark. Stage One: Only anomaly video fine-tuning. Stage Two: Anomaly video fine-tuning + Fine-tuning with SETS (Best results are shown in \textbf{bold}).}
\label{tab:ablation}
\vspace{-0.3cm}
\end{table*}

\textbf{Results on In-domain Dataset.} We first evaluate our method on the in-domain dataset, where the test set belongs to the same style and recording mode as the data used for training in Section~\ref{subsec:training}. As shown in Table~\ref{table:videobenchresults}, compared with the baseline~\cite{li2023llamavid}, with fewer visual embedding tokens and temporal effective tokens, our method brings more than double the performance improvement on total accuracy, also brings a significant increase in temporal localization. Driven by our proposed training strategy and designed effective tokens, more pure and effective visual-semantic information of abnormal events is efficiently aligned with LLMs and exhibits powerful anomaly video understanding capabilities. At the same time, we conduct fair comparisons with existing video understanding models \cite{videochat,videochatgpt,otter,valley,mplug,videollama} (see Table~\ref{table:videobenchresults}), and we demonstrate competitive performance. It is worth noting that we use the fewest tokens among all methods to achieve the state-of-the-art results on both total and temporal accuracies.

\noindent \textbf{Results on Cross-domain Dataset.} For evaluating the robustness and generalization of the models, we additionally design a cross-domain benchmark. We conduct a fair comparison of our method with the baseline~\cite{li2023llamavid} and the existing in-domain methods on the proposed cross-domain benchmark. The results presented in Table~\ref{table:videobenchresults} showcase a substantial performance improvement over existing methods on the cross-domain dataset, underscoring the exceptional generalization and temporal localization capabilities of our methodology. This clear superiority in performance serves as a compelling validation of the robustness and adaptability of our approach across diverse domains.

\noindent \textbf{Interaction with the Model.} We take some interactions with our well-trained model for better evaluation. As shown in Figure~\ref{fig:interaction}, we demonstrate the performance of our model in addressing various video anomaly understanding challenges. To evaluate the model's effectiveness, we select videos of different durations: short (0 to 1 minute), medium (1 to 30 minutes), and long (over 30 minutes). This variety allows us to thoroughly assess the model's capabilities in handling diverse anomaly understanding scenarios. In the Road Accident video (left side in Figure~\ref{fig:interaction}), our method successfully identifies a car driving at high speed and detects people falling, even in low-resolution footage. For the Explosion video (middle in Figure~\ref{fig:interaction}), the model accurately predicts the scene and the anomaly in a medium-length video depicting an explosion. In a normal video exceeding 30 minutes (right side in Figure~\ref{fig:interaction}), we demonstrate the model's ability to focus on both global and local information by asking it to summarize the content.

\noindent \textbf{Comparison on Other Benchmark.} We additionally compare our method on another benchmark (MMEval~\cite{du2024ECVA}) about anomaly video understanding with LLMs from different aspects. We follow a fair evaluation on our proposed model and obtain the quantity results as shown in Table~\ref{table:mmeval}, which shows the superiority of our method.

\subsection{Ablation Studies}

We conduct extensive ablation studies to validate the effectiveness of the key components in our method: Spatial Effective Token Selection (SETS, Section~\ref{subsec:SETS}) and Temporal Effective Tokens Generation (TETG, Section~\ref{subsec:TETG}) on progressive training strategies in Section~\ref{subsec:training}.

\noindent \textbf{Fine-tuning Stages.} The utilization of our high-quality UCF instruct-following data has proven to enhance the model's performance. Fine-tuning with this dataset has effectively contributed to a notable accuracy compared with the baseline. As evidenced in Table \ref{tab:ablation}, with our model designing (both SETS and TETG), the total accuracy for anomaly detection only achieves 25.12\% without any fine-tuning (denoted as Baseline). With anomaly video fine-tuning (denoted as Stage One Fine-tuning), the accuracy increases to 27.5\%. Furthermore, an efficient tuning with SETS (denoted as Stage Two Fine-tuning) can achieve our final performance of 30.69\% total accuracy. Temporal accuracy also shows similar increasing patterns with the scaling tuning stages.

\noindent \textbf{Effectiveness of Fine-tuning Data.} For a fair comparison, we fine-tune some high-performance comparison models \cite{videochatgpt,li2023llamavid} and compare them with our proposed
UCF instruct-following data. As shown in Table \ref{table:videobenchresultsinstru}, the performance of these comparison models has increased after fine-tuning, which proves the effectiveness of our proposed data. However, their performance is still not as good as our method, which proves the effectiveness of our proposed model.

\begin{table}[t]
\centering
\renewcommand{\arraystretch}{1.1}
\begin{center}
\resizebox{\linewidth}{!}{
\begin{tabular}{lccc}
\toprule\noalign{\smallskip}
\textbf{Method} & \textbf{LLM} & \textbf{Fine-tuned} {   } & \textbf{Total Acc.(\%)}$\uparrow$ \\
\midrule
\multirow{2}{*}{Video-ChatGPT} & \multirow{2}{*}{Vicuna-7B}{   } & - & 24.13 \\
& & \cellcolor{mygray}\checkmark & \cellcolor{mygray}26.23 \\
\midrule
\multirow{2}{*}{LLaMA-VID (\emph{Baseline}){  }} & \multirow{2}{*}{Vicuna-7B} & - & 14.83\\
& & \cellcolor{mygray}\checkmark & \cellcolor{mygray}23.10 \\
\midrule
\multirow{2}{*}{\emph{Ours}} & \multirow{2}{*}{Vicuna-7B} & - & 25.12 \\
& & \cellcolor{mygray}\checkmark & \cellcolor{mygray}\textbf{30.69} \\
\bottomrule
\end{tabular}}
\vspace{-0.5cm}
\end{center}
\caption{We fine-tune comparison models with our proposed UCF instruct-following data and evaluate the performance of these models before and after fine-tuning on the UCF benchmark.}
\vspace{-0.4cm}
\label{table:videobenchresultsinstru}
\end{table}
\setlength{\tabcolsep}{1pt}

\begin{figure*}[t]
\centering
\includegraphics[width=0.99\linewidth]{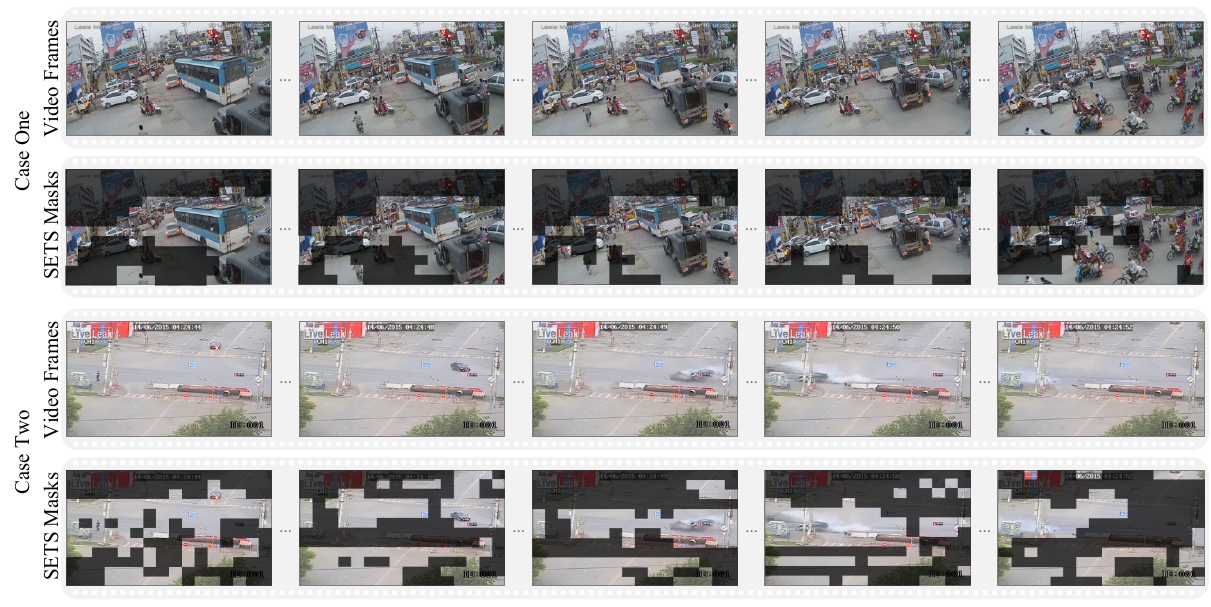}
\vspace{-0.1cm}
\caption{Visualization of the initial videos and our masked results. These two cases illustrate road accident scenarios: one occurring in a bustling street and the other in an empty suburb. Our SETS effectively filters redundant and irrelevant regions (with black patch-level masks).} 
\label{fig:mask}
\vspace{-0.3cm}
\end{figure*}

\noindent \textbf{Effectiveness of SETS.} Our proposed SETS demonstrates efficiency in extracting useful abnormal information, leading to performance enhancement. As illustrated in Table~\ref{tab:ablation}, with the SETS, the accuracy reaches 24.83\%, 25.86\% and 29.31\% without fine-tuning, anomaly video fine-tuning and fine-tuning with SETS, respectively, which far exceed the accuracy of the baseline. Its intuitive mechanism for information filtering can be further analyzed with reference to Figure~\ref{fig:mask}. The initial video is often cluttered with irrelevant and deceptive data. For example, in Figure~\ref{fig:mask}, case one illustrates a scenario where the overall structure is complex, yet only a small segment requires attention. The SETS effectively filters out the dynamic features that do not require attention. Similarly, as in case two, the abnormal area is quite small.
Our proposed SETS mechanism effectively filters out redundant and irrelevant information, significantly enhancing the model's ability to accurately pinpoint and recognize abnormal situations. 

We also conduct the ablation studies on $K$ ratios about SETS as shown in Tabel~\ref{table:kvalue}. Too small or too large $K$ ratios will cause performance degradation in both total and temporal accuracy. If $K$ ratios is too small, redundant information will affect the effectiveness of aligning abnormal event information with corresponding captions. In contrast, if $K$ ratios is too large, some important areas will be filtered out, resulting in information loss and suboptimal performance.



\noindent \textbf{Effectiveness of TETG.} Our proposed TETG generates tokens directly in the text token space of LLMs as priors for each video, offering robust priors on the temporal aspects of abnormal events. The provision results in performance improvement without the necessity for fine-tuning. As shown in Table \ref{tab:ablation}, the accuracy rises from 14.83\% to 23.79\%. Fine-tuning with anomaly video and SETS, the results even achieve 26.10\% and 30.69\% independently, which manifests the effectiveness of TETG. Besides, the integration of SETS and TETG highlights the importance of leveraging spatial and temporal information effectively in anomaly detection systems, boosting the results to 25.15\%, 27.50\% and 30.69\%, respectively.

\begin{table}[t]
\small
\centering
\renewcommand{\arraystretch}{1.1}
\resizebox{1.0\linewidth}{!}{ 
\begin{tabular}{lccccc}
    \toprule
    \textbf{\# Sample Rate $K$} {   } & \textbf{0.1} & \textbf{0.3} & \cellcolor{mygray}\textbf{0.5} & \textbf{0.7} & \textbf{0.9}\\
    \midrule
    \textbf{Total Acc.(\%)}$\uparrow$ & {   } 23.61 {   } & {   } 24.83 {   } & \cellcolor{mygray} {   } 30.69 {   } & {   } 28.67 {   } & {   } 27.27 {   } \\
    \textbf{Temporal Acc.(\%)}$\uparrow$ & {   } 29.03 {   } & {   } 29.93 {   } & \cellcolor{mygray} {   } 35.00 {   } & {   } 31.23 {   } & {   } 31.03 {   } \\
    \bottomrule
\end{tabular}}
\vspace{-0.1cm}
\caption{For the sample rate of tokens, $K$ ratios in SETS, we sample the patch tokens with ordered distance at a fixed sample rate (0.5). The ablation indicates too large sampling rates cause too much context noise, and too small sampling rates lose visual information.}
\vspace{-0.7cm}
\label{table:kvalue}
\end{table}


 
\section{Discussion}

\textbf{Key tokens play key roles.} To the best of our knowledge, we are the first to explore how to assign different learnable knowledge to different tokens for better alignment with LLMs on visual contents, thus promoting video anomaly detection and understanding (see Table~\ref{table:videobenchresults} and Figure~\ref{fig:interaction}). We assign the most effective roles to different tokens in both spatial and temporal dimensions, enabling the model to handle various tokens more efficiently. The video contains abundant but redundant information. Our proposed SETS and TETG effectively compress the spatial and temporal information of abnormal events respectively, and utilize the existing alignment mechanism of MLLMs at a very low cost to participate in LLMs' reasoning (see Table~\ref{tab:ablation}). Our exploration inspires more representation learning of MLLMs to facilitate downstream tasks.

\noindent \textbf{Data matters a lot.} We construct instruct-following data for anomaly videos, containing approximately 4,000 videos, which is significantly less than the amount of baseline fine-tuning video data (\eg, over 90k videos for fine-tuning in baseline model~\cite{li2023llamavid}). We still achieve promising performance on both in-domain and cross-domain benchmarks (see Table~\ref{table:videobenchresults}). This relies on the high-quality Question-Answer pairs in our instruct-following data. Meanwhile, SETS also improves data quality during fine-tuning: visual areas irrelevant to Question-Answer pairs are filtered out (see Figure~\ref{fig:mask}), which allows for significant performance improvements in the second stage of fine-tuning (see Section~\ref{subsec:training}) with very few steps (less than 150 iterations).

\noindent \textbf{Broader impacts.} Video anomaly understanding has far-reaching implications across various sectors, including security, healthcare, industrial safety, and so on. By enhancing the ability to automatically identify and respond to unusual or suspicious activities in real-time, LLMs can significantly improve public safety, crime prevention, patient monitoring, hazard detection, loss prevention, traffic management, and urban planning. 
These systems offer substantial benefits in terms of operational efficiency and safety.

\noindent \textbf{Limitations.} Although our model adeptly portrays the occurrence, type, and area of video abnormal events, it still faces challenges in detecting and describing certain complex scenes. Our strategy represents an early successful validation and investigation of large models for video anomaly identification and localization. Consequently, our method possesses significant potential for enhancement in recognizing diverse abnormal video scenes. These insights motivate us to continue pursuing more powerful and efficient video anomaly understanding technologies in the future, aiming to address more challenges in the real world~\cite{lyu2024total, liu2023mars3d,zhang2023polar}.

\section{Conclusions}

In this paper, we propose a novel MLLM for understanding anomalies in videos with LLMs by aligning effective tokens in both temporal and spatial space. The proposed method includes Spatial Effective Token Selection (SETS) for identifying abnormal events in small areas of large scenes and Temporal Effective Tokens Generation (TETG) for addressing the sparseness of abnormal events in video time sequences. We also develop instruct-following data of video anomaly detection to fine-tune the model. Besides, evaluation on the video anomaly understanding benchmark and a proposed cross-domain benchmark demonstrates the effectiveness of the proposed method. It further presents a promising approach for video anomaly understanding using MLLMs, showcasing the potential of effective tokens for enhancing video understanding tasks.

\clearpage

\section*{Acknowledgements}

This work has been supported in part by Hong Kong Research Grant Council - Early Career Scheme (Grant No.27209621), General Research Fund Scheme (Grant No.17202422, 17212923, 17215025), Theme-based Research (Grant No.T45-701/22-R) and Shenzhen Science and Technology Innovation Commission (SGDX20220530111405040). Part of the described research work is conducted in the JC STEM Lab of Robotics for Soft Materials funded by The Hong Kong Jockey Club Charities Trust.
{
    \small
    \bibliographystyle{ieeenat_fullname}
    \bibliography{main}
}


\end{document}